\documentclass[11pt,a4paper]{article}
\usepackage[hyperref]{acl2019}
\usepackage{times}
\usepackage{latexsym}
\usepackage{CJKutf8}
\usepackage{booktabs}
\usepackage{multirow}
\usepackage{amsmath}
\usepackage{amssymb}
\usepackage{url}
\usepackage{graphicx}
\usepackage{helvet}  
\usepackage{courier}
\usepackage{threeparttable}
\usepackage{amsthm}
\usepackage{CJKutf8}

\usepackage{algorithm}
\usepackage{algpseudocode}
\usepackage{amsmath}
\usepackage{graphics}
\usepackage{epsfig}
\theoremstyle{definition}

\usepackage{mydef}

\aclfinalcopy
\title{How to Fine-Tune BERT for Text Classification?}

\author{Chi Sun, Xipeng Qiu\thanks{\ \  Corresponding author} , Yige Xu, Xuanjing Huang\\
	Shanghai Key Laboratory of Intelligent Information Processing, Fudan University\\
	School of Computer Science, Fudan University\\
	825 Zhangheng Road, Shanghai, China\\
	{\tt \{sunc17,xpqiu,ygxu18,xjhuang\}@fudan.edu.cn} \\}

\date{}

\begin{document}
	\begin{CJK*}{UTF8}{gbsn}
	\maketitle
	\begin{abstract}
		Language model pre-training has proven to be useful in learning universal language representations. As a state-of-the-art language model pre-training model, BERT (Bidirectional Encoder Representations from Transformers) has achieved amazing results in many language understanding tasks. In this paper, we conduct exhaustive experiments to investigate different fine-tuning methods of BERT on text classification task and provide a general solution for BERT fine-tuning. Finally, the proposed solution obtains new state-of-the-art results on eight widely-studied text classification datasets.\footnote{The source codes are available at \url{https://github.com/xuyige/BERT4doc-Classification}.}
	\end{abstract}
	
	\section{Introduction}
    Text classification is a classic problem in Natural Language Processing (NLP). The task is to assign predefined categories to a given text sequence. An important intermediate step is the text representation. Previous work uses various neural models to learn text representation, including convolution models \cite{kalchbrenner2014convolutional,zhang2015character,conneau2016very,johnson2017deep,zhang2017deconvolutional,shen2018deconvolutional}, recurrent models  \cite{liu2016recurrent,yogatama2017generative,seo2017neural}, and attention mechanisms \cite{yang2016hierarchical,lin2017structured}.

	Alternatively, substantial work has shown that pre-trained models on large corpus are beneficial for text classification and other NLP tasks, which can avoid training a new model from scratch.
One kind of pre-trained models is the word embeddings, such as word2vec \cite{mikolov2013distributed} and GloVe \cite{pennington2014glove}, or the contextualized word embeddings, such as CoVe \cite{mccann2017learned} and ELMo 
\cite{peters2018deep}. These word embeddings are often used as additional features for the main task.
Another kind of pre-training models is sentence-level. \citet{howard2018universal} propose ULMFiT, a fine-tuning method for pre-trained language model that achieves state-of-the-art results on six widely studied text classification datasets. More recently, pre-trained language models have shown to be useful in learning common language representations by utilizing a large amount of unlabeled data: e.g., OpenAI GPT \cite{radford2018improving} and BERT \cite{devlin2018bert}. BERT is based on a multi-layer bidirectional Transformer \cite{vaswani2017attention} and is trained on plain text for masked word prediction and next sentence prediction tasks.
	
Although BERT has achieved amazing results in many natural language understanding (NLU) tasks, its potential has yet to be fully explored. There is little research to enhance BERT to improve the performance on target tasks further.

In this paper, we investigate how to maximize the utilization of BERT for the text classification task. We explore several ways of fine-tuning BERT to enhance its performance on text classification task. We design exhaustive experiments to make a detailed analysis of BERT.

The contributions of our paper are as follows:
\begin{itemize*}
	\item
	We propose a general solution to fine-tune the pre-trained BERT model, which includes three steps: (1) further pre-train BERT on within-task training data or in-domain data; (2) optional fine-tuning BERT with multi-task learning if several related tasks are available; (3) fine-tune BERT for the target task.
	\item We also investigate the fine-tuning methods for BERT on target task, including pre-process of long text, layer selection, layer-wise learning rate, catastrophic forgetting, and low-shot learning problems.
	\item
	We achieve the new state-of-the-art results on seven widely-studied English text classification datasets and one Chinese news classification dataset.
	
\end{itemize*}
	
\section{Related Work}
 Borrowing the learned knowledge from the other tasks has a rising interest in the field of NLP. We briefly review two related approaches: language model pre-training and multi-task Learning.

\subsection{Language Model Pre-training}

	Pre-trained word embeddings \cite{mikolov2013distributed,pennington2014glove}, as an important component of modern NLP systems can offer significant improvements over embeddings learned from scratch. The generalization of word embeddings, such as sentence embeddings \cite{kiros2015skip,logeswaran2018efficient} or paragraph embeddings \cite{le2014distributed}, are also used as features in downstream models.
	
	\citet{peters2018deep} concatenate embeddings derived from language model as additional features for the main task and advance the state-of-the-art for several major NLP benchmarks. In addition to pre-training with unsupervised data, transfer learning with a large amount of supervised data can also achieve good performance, such as natural language inference \cite{conneau2017supervised} and machine translation \cite{mccann2017learned}.
	
	More recently, the method of pre-training language models on a large network with a large amount of unlabeled data and fine-tuning in downstream tasks has made a breakthrough in several natural language understanding tasks, such as OpenAI GPT \cite{radford2018improving} and BERT \cite{devlin2018bert}. \citet{dai2015semi} use language model fine-tuning but overfit with 10k labeled examples while \citet{howard2018universal} propose ULMFiT and achieve state-of-the-art results in the text classification task. BERT is pre-trained on \textit{Masked Language Model Task} and \textit{Next Sentence Prediction Task} via a large cross-domain corpus. Unlike previous bidirectional language models (biLM) limited to a combination of two unidirectional language models (i.e., left-to-right and right-to-left), BERT uses a Masked Language Model to predict words which are randomly masked or replaced. BERT is the first fine-tuning based representation model that achieves state-of-the-art results for a range of NLP tasks, demonstrating the enormous potential of the fine-tuning method. In this paper, we have further explored the BERT fine-tuning method for text classification.

	\subsection{Multi-task learning}
	
	Multi-task learning \cite{richcaruana1993,collobert2008unified} is another relevant direction. \citet{rei2017semi} and \citet{liu2018empower} use this method to train the language model and the main task model jointly. \citet{liu2019multi} extend the MT-DNN model originally proposed in \citet{liu2015representation} by incorporating BERT as its shared text encoding layers. MTL requires training tasks from scratch every time, which makes it inefficient and it usually requires careful weighing of task-specific objective functions \cite{chen2017gradnorm}. However, we can use multi-task BERT fine-tuning to avoid this problem by making full use of the shared pre-trained model.

\section{BERT for Text Classification}

BERT-base model contains an encoder with 12 Transformer blocks, 12 self-attention heads, and the hidden size of 768. BERT takes an input of a sequence of no more than 512 tokens and outputs the representation of the sequence. The sequence has one or two segments that the first token of the sequence is always \texttt{[CLS]} which contains the special classification embedding and another special token \texttt{[SEP]} is used for separating segments.
	
For text classification tasks, BERT takes the final hidden state $\bh$ of the first token \texttt{[CLS]} as the representation of the whole sequence. A simple softmax classifier is added to the top of BERT to predict the probability of label $c$:
\begin{align}
p(c|\bh) =\mathrm{softmax}(W\bh),
\end{align}
where $W$ is the task-specific parameter matrix. We fine-tune all the parameters from BERT as well as $W$ jointly by maximizing the log-probability of the correct label.
	
\section{Methodology}

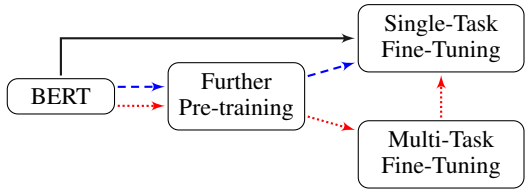
\begin{figure}
  \centering

  \tikzstyle{edge}=[-latex',thick,draw=black!90,shorten <=1pt,shorten >=1pt]
\tikzstyle{block}=[draw, text width=5em,align=center,shape=rectangle, rounded corners, , align=center]
  \tikzstyle{nobox}=[align=center]

\begin{tikzpicture}[node distance=6em,auto,font=\footnotesize\selectfont]

    \node [block,text width=3em,] (bert) {BERT};
    \node [block, text width=4em,] (x1) [right of=bert] {Further Pre-training};

    \node [block] (x2) [right of=x1,node distance=7em,yshift=2em]{Single-Task Fine-Tuning};
    \node [block] (x3) [right of=x1,node distance=7em,,yshift=-2em] {Multi-Task Fine-Tuning};

    \path[edge,] (bert) |-  (x2);

    \path[edge,densely dashed,,blue,] (bert.10) --  (x1.172);
    \path[edge,densely dashed,,blue,] (x1) -- (x2);

    \path[edge,densely dotted,red] (x1) -- (x3);
    \path[edge,red,densely dotted] (x3) -- (x2);
    \path[edge,red,densely dotted] (bert.-10) --  (x1.-172);
\end{tikzpicture}
	  \caption{Three general ways for fine-tuning BERT, shown with different colors.}\label{fig:arch}
\end{figure}

When we adapt BERT to NLP tasks in a target domain, a proper fine-tuning strategy is desired. In this paper, we look for the proper fine-tuning methods in the following three ways.

1) \textbf{Fine-Tuning Strategies}: When we fine-tune BERT for a target task, there are many ways to utilize BERT. For example, the different layers of BERT capture different levels of semantic and syntactic information, which layer is better for a target task? How we choose a better optimization algorithm and learning rate?

2) \textbf{Further Pre-training}: BERT is trained in the general domain, which has a different data distribution from the target domain. A natural idea is to further pre-train BERT with target domain data.

3) \textbf{Multi-Task Fine-Tuning}: Without pre-trained LM models, multi-task learning has shown its effectiveness of exploiting the shared knowledge among the multiple tasks. When there are several available tasks in a target domain, an interesting question is whether it still bring benefits to fine-tune BERT on all the tasks simultaneously.

Our general methodology of fine-tuning BERT is shown in Figure \ref{fig:arch}.

\subsection{Fine-Tuning Strategies}

Different layers of a neural network can capture different levels of syntactic and semantic information \cite{yosinski2014transferable,howard2018universal}.

To adapt BERT to a target task, we need to consider several factors: 1) The first factor is the preprocessing of long text since the maximum sequence length of BERT is 512. 2) The second factor is layer selection. The official BERT-base model consists of an embedding layer, a 12-layer encoder, and a pooling layer. We need to select the most effective layer for the text classification task. 3) The third factor is the overfitting problem. A better optimizer with an appropriate learning rate is desired.

Intuitively, the lower layer of the BERT model may contain more general information. We can fine-tune them with different learning rates.

	

Following \citet{howard2018universal}, we split the parameters $\theta$ into $\{\theta^1,\cdots,\theta^L\}$ where $\theta^l$ contains the parameters of the $l$-th layer of BERT. Then the parameters are updated as follows:
	\begin{align}	\theta^l_{t}=\theta^l_{t-1}-\eta^l\cdot\nabla_{\theta^l}J(\theta),
\label{eq:sgd}
	\end{align}
where $\eta^l$ represents the learning rate of the $l$-th layer.

We set the base learning rate to $\eta^L$ and use $\eta^{k-1}=\xi \cdot\eta^{k}$, where $\xi$ is a decay factor and less than or equal to 1.
When $\xi<1$, the lower layer has a lower learning rate than the higher layer. When $\xi=1$, all layers have the same learning rate, which is equivalent to the regular stochastic gradient descent (SGD). We will investigate these factors in Sec. \ref{sec:exp-fit}.

\subsection{Further Pre-training}

The BERT model is pre-trained in the general-domain corpus. For a text classification task in a specific domain, such as movie reviews, its data distribution may be different from BERT. Therefore, we can further pre-train BERT with masked language model and next sentence prediction tasks on the domain-specific data. Three further pre-training approaches are performed:

1) Within-task pre-training, in which BERT is further pre-trained on the training data of a target task.

2) In-domain pre-training, in which the pre-training data is obtained from the same domain of a target task. For example, there are several different sentiment classification tasks, which have a similar data distribution. We can further pre-train BERT on the combined training data from these tasks.

3) Cross-domain pre-training, in which the pre-training data is obtained from both the same and other different domains to a target task.
	


%

We will investigate these different approaches to further pre-training in Sec. \ref{sec:exp-pretrain}.
	
\begin{table*}[t!]\small\setlength{\tabcolsep}{8pt}
    \centering
    \begin{tabular}{c c c c c c c c}
    \toprule
    \multirow{2}*{Dataset} & \multirow{2}*{Classes} & \multirow{2}*{Type} & Average & Max & Exceeding & Train & Test \\
    ~ & ~ & ~ & lengths & lengths & ratio & samples & samples \\
    \midrule

    IMDb & 2 & Sentiment & 292 & 3,045 & 12.69\% & 25,000 & 25,000  \\
    Yelp P. & 2 & Sentiment & 177 & 2,066 & 4.60\% & 560,000 & 38,000 \\
    Yelp F. & 5 & Sentiment & 179 & 2,342 & 4.60\% & 650,000 & 50,000 \\
    TREC & 6 & Question & 11 & 39 & 0.00\% & 5,452 & 500 \\
    Yahoo! Answers & 10 & Question & 131 & 4,018 & 2.65\% & 1,400,000 & 60,000 \\
    AG's News & 4 & Topic & 44 & 221 & 0.00\% & 120,000 & 7,600  \\
    DBPedia & 14 & Topic & 67 & 3,841 & 0.00\% & 560,000 & 70,000 \\
    Sogou News & 6 & Topic & 737 & 47,988 & 46.23\% & 54,000 & 6,000  \\
    \bottomrule
\end{tabular}
		\caption{\label{table_1} Statistics of eight text classification datasets. The exceeding ratio means the percentage of the number of samples with a length exceeding 512.
		}
	\end{table*}

\subsection{Multi-Task Fine-Tuning} \label{multi}
	

Multi-task Learning is also an effective approach to share the knowledge obtained from several related supervised tasks. Similar to \citet{liu2019multi}, we also use fine-tune BERT in multi-task learning framework for text classification.

All the tasks share the BERT layers and the embedding layer. The only layer that does not share is the final classification layer, which means that each task has a private classifier layer. The experimental analysis is in Sec. \ref{sec:exp-mtl}.


	\section{Experiments}
    We investigate the different fine-tuning methods for seven English and one Chinese text classification tasks. We use the base BERT models: the uncased BERT-base model\footnote{https://storage.googleapis.com/bert\_models/2018\_10\_18/\\uncased\_L-12\_H-768\_A-12.zip} and the Chinese BERT-base model\footnote{https://storage.googleapis.com/bert\_models/2018\_11\_03/\\chinese\_L-12\_H-768\_A-12.zip} respectively.

	\subsection{Datasets} \label{s_1}
	We evaluate our approach on eight widely-studied datasets. These datasets have varying numbers of documents and varying document lengths, covering three common text classification tasks: sentiment analysis, question classification, and topic classification. We show the statistics for each dataset in Table \ref{table_1}.
	
	\paragraph{Sentiment analysis}
	\quad For sentiment analysis, we use the binary film review IMDb dataset \cite{maas2011learning} and the binary and five-class version of the Yelp review dataset built by \citet{zhang2015character}.
	
	{\noindent \textbf{Question classification}}
	\quad For question classification, we evaluate our method on the six-class version of the TREC dataset \cite{voorhees1999trec} and Yahoo! Answers dataset created by \citet{zhang2015character}. TREC dataset is dataset for question classification consisting of open-domain, fact-based questions divided into broad semantic categories. Compared to other document-level datasets, TREC dataset is sentence-level, and there are fewer training examples for it. Yahoo! Answers dataset is a big dataset with 1,400k train samples.
	
    {\noindent \textbf{Topic classification}}
	\quad For topic classification, we use large-scale AG's News and DBPedia created by \citet{zhang2015character}. To test the effectiveness of BERT for Chinese text, we create the Chinese training and test datasets for Sogou news corpus. Unlike \citet{zhang2015character}, we use the Chinese character directly rather than Pinyin. The dataset is a combination of the SogouCA and SogouCS news corpora \cite{wang2008automatic}. We determine the category of the news based on the URL, such as ``sports" corresponding to ``http://sports.sohu.com". We choose 6 categories – ``sports", ``house", ``business”, ``entertainment”, ``women" and ``technology”. The number of training samples selected for each class is 9,000 and testing 1,000.
	
	{\noindent \textbf{Data preprocessing}}
	\quad Following \citet{devlin2018bert}, we use WordPiece embeddings \cite{wu2016google} with a 30,000 token vocabulary and denote split word pieces with \#\#. So the statistics of the length of the documents in the datasets are based on the word pieces. For further pre-training with BERT, we use spaCy\footnote{https://spacy.io/} to perform sentence segmentation in English datasets and we use ``。",``？" and ``！" as separators when dealing with the Chinese Sogou News dataset.
	
	\subsection{Hyperparameters}
	We use the BERT-base model \cite{devlin2018bert} with a hidden size of 768, 12 Transformer blocks \cite{vaswani2017attention} and 12 self-attention heads. We further pre-train with BERT on 1 TITAN Xp GPU, with a batch size of 32, max squence length of 128, learning rate of 5e-5, train steps of 100,000 and warm-up steps of 10,000.
	
	We fine-tune the BERT model on 4 TITAN Xp GPUs and set the batch size to 24 to ensure that the GPU memory is fully utilized. The dropout probability is always kept at 0.1. We use Adam with $\beta_1=0.9$ and $\beta_2=0.999$. We use \textit{slanted triangular learning rates} \cite{howard2018universal}, the base learning rate is 2e-5, and the warm-up proportion is 0.1. We empirically set the max number of the epoch to 4 and save the best model on the validation set for testing.

\subsection{Exp-I: Investigating Different Fine-Tuning Strategies} \label{sec:exp-fit}

In this subsection, we use the IMDb dataset to investigate the different fine-tuning strategies. The official pre-trained model is set as the initial encoder\footnote{https://github.com/google-research/bert}.

\subsubsection{Dealing with long texts}

The maximum sequence length of BERT is 512. The first problem of applying BERT to text classification is how processing the text with a length larger than 512. We try the following ways for dealing with long articles.

\paragraph{Truncation methods}
Usually, the key information of an article is at the beginning and end. We use three different methods of truncate text to perform BERT fine-tuning.
\begin{enumerate**}
	\item \textbf{head-only}: keep the first 510 tokens\footnote{512 to subtract the \texttt{[CLS]} and \texttt{[SEP]} tokens.};
	\item \textbf{tail-only}: keep the last 510 tokens;
	\item \textbf{head+tail}: empirically select the first 128 and the last 382 tokens.
\end{enumerate**}

\paragraph{Hierarchical methods}
The input text is firstly divided into $k=L/510$ fractions, which is fed into BERT to obtain the representation of the $k$ text fractions. The representation of each fraction is the hidden state of the \texttt{[CLS]} tokens of the last layer. Then we use mean pooling, max pooling and self-attention to combine the representations of all the fractions.

Table \ref{tb:longtext} shows the effectiveness of the above methods. The truncation method of \textbf{head+tail} achieves the best performance on IMDb and Sogou datasets. Therefore, we use this method to deal with the long text in the following experiments.

	\begin{table}[ht!]\small\setlength{\tabcolsep}{8pt}
		\centering
		\begin{tabular}{l c c}
			\toprule
			Method & IMDb & Sogou \\
			\midrule
			head-only & 5.63 & 2.58 \\
			tail-only & 5.44 & 3.17 \\
			head+tail & \textbf{5.42} & \textbf{2.43} \\
			hier. mean & 5.89 & 2.83  \\
			hier. max & 5.71 & 2.47 \\
			hier. self-attention & 5.49 & 2.65 \\			
			\bottomrule
		\end{tabular}
		\caption{Test error rates (\%) on IMDb and Chinese Sogou News datasets.
		}\label{tb:longtext}
	\end{table}	

\begin{figure*}[h!]
\centering
\subfloat[lr=2e-5]{
\includegraphics[width=0.25\linewidth]{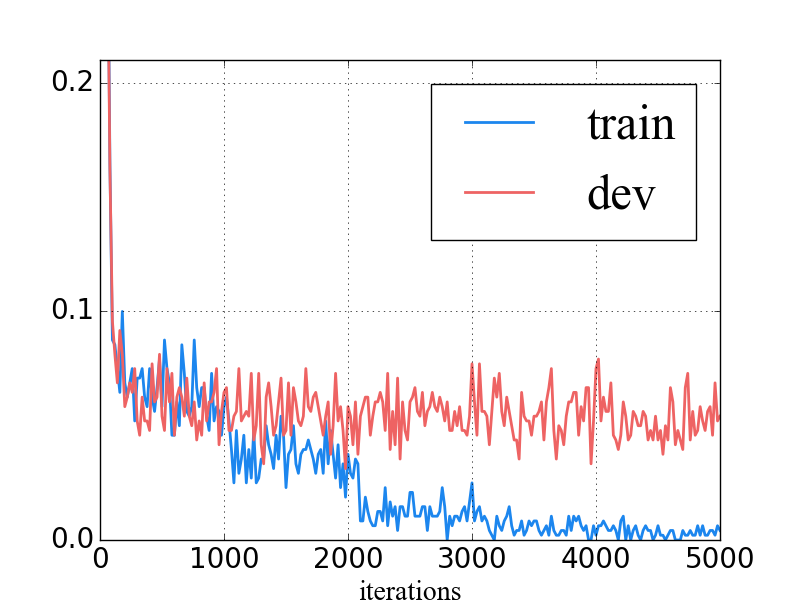}
}
\subfloat[lr=5e-5]{
\includegraphics[width=0.25\linewidth]{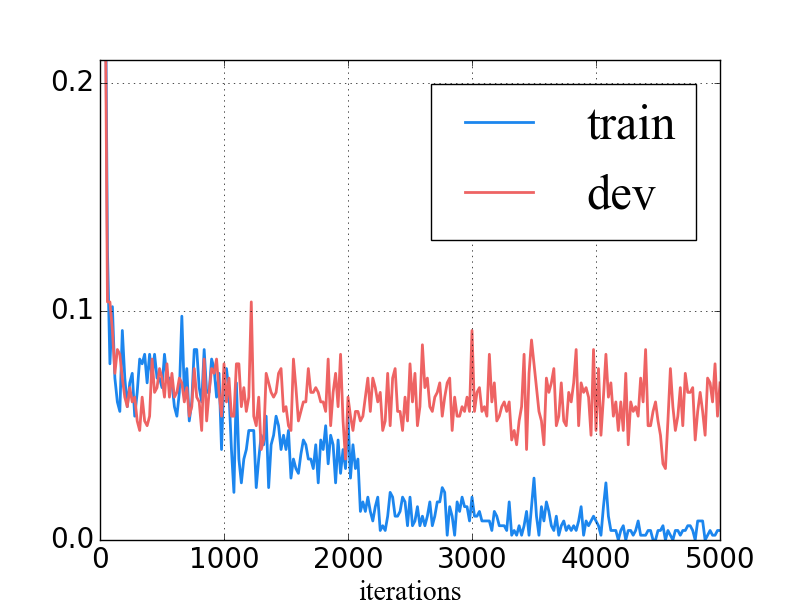}
}
\subfloat[lr=1e-4]{
\includegraphics[width=0.25\linewidth]{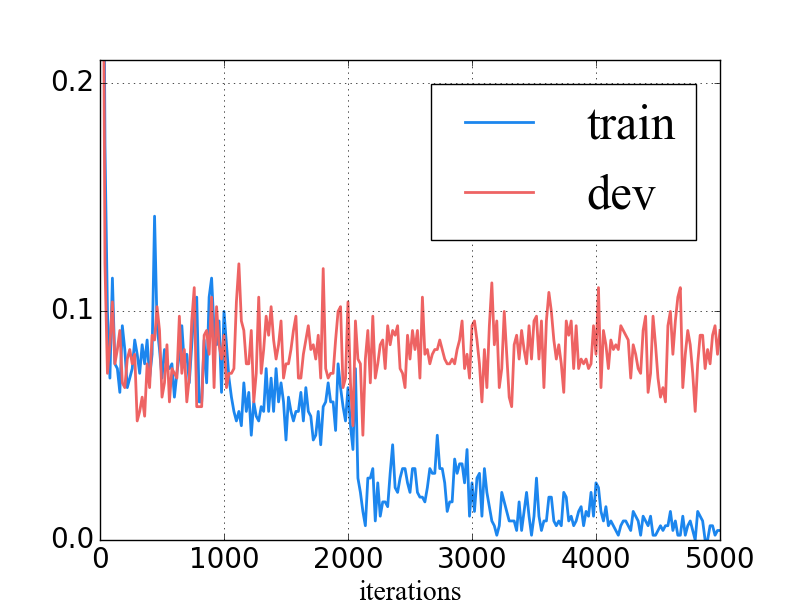}
}
\subfloat[lr=4e-4]{
	\includegraphics[width=0.25\linewidth]{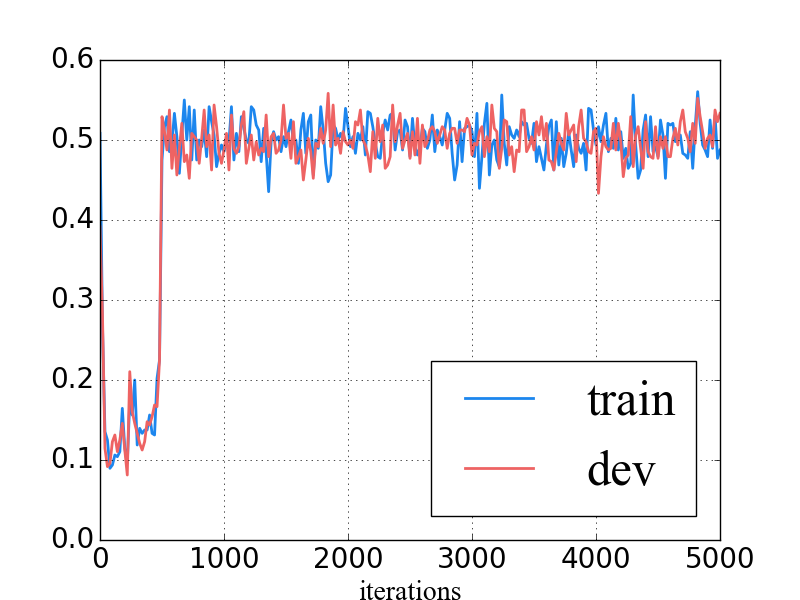}
}
	\caption{Catastrophic Forgetting}\label{fig:cat-forget}
\end{figure*}

\subsubsection{Features from Different layers}
Each layer of BERT captures the different features of the input text. We investigate the effectiveness of features from different layers. We then fine-tune the model and record the performance on test error rates.

Table \ref{tb:layers} shows the performance of fine-tuning BERT with different layers. The feature from the last layer of BERT gives the best performance. Therefore, we use this setting for the following experiments.


\begin{table}[h!]\small
\centering
\begin{tabular}{l c}
	\toprule
	Layer & Test error rates(\%) \\
	\midrule
	Layer-0 & 11.07 \\
	Layer-1 & 9.81 \\
	Layer-2 & 9.29 \\
	Layer-3 & 8.66 \\
	Layer-4 & 7.83 \\
	Layer-5 & 6.83 \\
	Layer-6 & 6.83 \\
	Layer-7 & 6.41 \\
	Layer-8 & 6.04 \\
	Layer-9 & 5.70 \\
	Layer-10 & 5.46 \\
	Layer-11 & \textbf{5.42} \\
	\midrule
	First 4 Layers + concat & 8.69 \\
	First 4 Layers + mean & 9.09 \\
	First 4 Layers + max & 8.76 \\
	\midrule
	Last 4 Layers + concat & 5.43 \\
	Last 4 Layers + mean & 5.44 \\
	Last 4 Layers + max & \textbf{5.42} \\
	\midrule
	All 12 Layers + concat & 5.44 \\
	\bottomrule
\end{tabular}
\caption{Fine-tuning BERT with different layers on IMDb dataset.
}\label{tb:layers}
\end{table}


\subsubsection{Catastrophic Forgetting}
Catastrophic forgetting \cite{mccloskey1989catastrophic} is usually a common problem in transfer learning, which means the pre-trained knowledge is erased during learning of new knowledge. Therefore, we also investigate whether BERT suffers from the catastrophic forgetting problem.

We fine-tune BERT with different learning rates, and the learning curves of error rates on IMDb are shown in Figure \ref{fig:cat-forget}.

We find that a lower learning rate, such as 2e-5, is necessary to make BERT overcome the catastrophic forgetting problem. With an aggressive learn rate of 4e-4, the training set fails to converge.

\subsubsection{Layer-wise Decreasing Layer Rate}
Table \ref{tb:layer-wise-lr} show the performance of different base learning rate and decay factors (see Eq. \eqref{eq:sgd}) on IMDb dataset. We find that assign a lower learning rate to the lower layer is effective to fine-tuning BERT, and an appropriate setting is $\xi$=0.95 and lr=2.0e-5.

\begin{table}[h!]\small
\centering
\begin{tabular}{c c c}
	\toprule
	Learning rate & Decay factor $\xi$ & Test error rates(\%) \\
	\midrule
	2.5e-5 & 1.00 & 5.52 \\
	2.5e-5 & 0.95 & 5.46 \\
    2.5e-5 & 0.90 & \textbf{5.44} \\
    2.5e-5 & 0.85 & 5.58 \\
    \midrule
    2.0e-5 & 1.00 & 5.42 \\
    2.0e-5 & 0.95 & \textbf{5.40} \\
    2.0e-5 & 0.90 & 5.52 \\
    2.0e-5 & 0.85 & 5.65 \\
	\bottomrule
\end{tabular}
\caption{Decreasing layer-wise layer rate.
}\label{tb:layer-wise-lr}
\end{table}



\subsection{Exp-II: Investigating the Further Pretraining}\label{sec:exp-pretrain}

Besides, fine-tune BERT with supervised learning, we can further pre-train BERT on the training data by unsupervised masked language model and next sentence prediction tasks.
In this section, we investigate the effectiveness of further pre-training. In the following experiments, we use the best strategies in Exp-I during the fine-tuning phase.

\begin{table*}[t!]\small\setlength{\tabcolsep}{10pt}
		\centering
		\begin{tabular}{l |c c c| c c| c c}
			\toprule
            Domain & \multicolumn{3}{c}{sentiment} & \multicolumn{2}{|c}{question} & \multicolumn{2}{|c}{topic}\\
			\midrule
			Dataset & IMDb & Yelp P. & Yelp F. & TREC & Yah. A. & AG's News & DBPedia  \\
			\midrule
			IMDb & \textbf{4.37} & 2.18 & 29.60 & 2.60 & 22.39 & 5.24 & 0.68 \\
			Yelp P. & 5.24 & 1.92 & 29.37 & 2.00 & 22.38 & 5.14 & \textbf{0.65} \\
			Yelp F. & 5.18 & 1.94 & 29.42 & 2.40 & 22.33 & 5.43 & \textbf{0.65}\\
			all sentiment & 4.88 & \textbf{1.87} & 29.25 & 3.00 & 22.35 & 5.34 & 0.67 \\
			\midrule
			TREC & 5.65 & 2.09 & 29.35 & 3.20 & 22.17 & 5.12 & 0.66 \\
			Yah. A. & 5.52 & 2.08 & 29.31 & \textbf{1.80} & 22.38 & 5.16 & 0.67 \\
			all question & 5.68 & 2.14 & 29.52 & 2.20 & \textbf{21.86} & 5.21 & 0.68 \\
			\midrule
			AG's News & 5.97 & 2.15 & 29.38 & 2.00 & 22.32 & \textbf{4.80} & 0.68 \\
			DBPedia & 5.80 & 2.13 & 29.47 & 2.60 & 22.30 & 5.13 & 0.68 \\
			all topic & 5.85 & 2.20 & 29.68 & 2.60 & 22.28 & 4.88 & \textbf{0.65} \\
			\midrule
			all & 5.18 & 1.97 & \textbf{29.20} & 2.80 & 21.94 & 5.08 & 0.67 \\
\midrule
w/o pretrain & 5.40 & 2.28 & 30.06 & 2.80 & 22.42 & 5.25 & 0.71\\
			
			\bottomrule
		\end{tabular}
		\caption{Performance of in-domain and cross-domain further pre-training on seven datasets. Each was further pre-trained for 100k steps. The first column indicates the different further pre-training dataset. ``all sentiment'' means the dataset consists of all the training datasets in sentiment domain. ``all'' means the dataset consists of all the seven training datasets. Note that some of the data in Yelp P. and Yelp F. are overlapping, e.g., part of the data in the test set of Yelp P. will appear in the training set of Yelp F., so we remove this part of data from the training sets during further pre-training.
		}\label{tb:in-cross}
	\end{table*}

\subsubsection{Within-Task Further Pre-Training}

 Therefore, we first investigate the effectiveness of within-task further pre-training. We take further pre-trained models with different steps and then fine-tune them with text classification task.

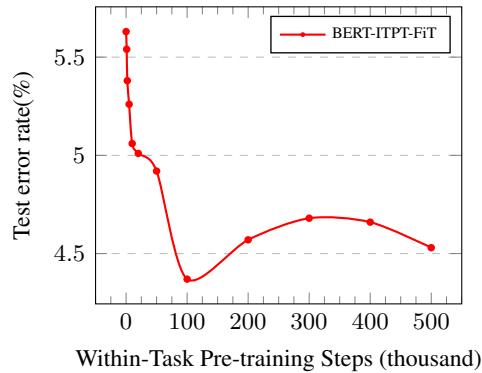
\begin{figure}[h!]
\centering
\pgfplotsset{width=0.4\textwidth}

\begin{tikzpicture}
\pgfplotstableread{./further-pretrain.txt}\datatable
  \begin{axis}[
font=\small,
xtick scale label code/.code={\pgfmathparse{int(-#1)}$x \cdot 10^{\pgfmathresult}$},
    every x tick scale label/.style={at={(xticklabel cs:0.5)}, anchor = north},
  xlabel={Within-Task Pre-training Steps (thousand)},
  minor x tick num=3,
  xtick distance=100,
  ylabel={Test error rate(\%)},
  legend entries={BERT-ITPT-FiT},
  mark size=1.0pt,
  ymajorgrids=true,
  grid style=dashed,
  legend pos= north east,
  legend style={font=\tiny,line width=.5pt,mark size=.5pt,
          /tikz/every even column/.append style={column sep=0.5em}},
          smooth,
  ]

  \addplot [red,thick,mark=*,
  ] table [x=step, y=BERT] {\datatable};

  \end{axis}
\end{tikzpicture}
\caption{Benefit of different further pre-training steps on IMDb datasets. BERT-ITPT-FiT means ``BERT + withIn-Task Pre-Training + Fine-Tuning''. }\label{fig:further-pretrain}
\end{figure}

As shown in Figure \ref{fig:further-pretrain}, the further pre-training is useful to improve the performance of BERT for a target task, which achieves the best performance after 100K training steps.

\subsubsection{In-Domain and Cross-Domain Further Pre-Training}

Besides the training data of a target task, we can further pre-train BERT on the data from the same domain.
In this subsection, we investigate whether further pre-training BERT with in-domain and cross-domain data can continue to improve the performance of BERT.

We partition the seven English datasets into three domains: topic, sentiment, and question. The partition way is not strictly correct. Therefore we also conduct extensive experiments for cross-task pre-training, in which each task is regarded as a different domain.

The results is shown in Table \ref{tb:in-cross}. We find that almost all further pre-training models perform better on all seven datasets than the original BERT-base model (row `w/o pretrain' in Table \ref{tb:in-cross}). Generally, in-domain pretraining can bring better performance than within-task pretraining. On the small sentence-level TREC dataset, within-task pre-training do harm to the performance while in-domain pre-training which utilizes Yah. A. corpus can achieve better results on TREC.

Cross-domain pre-training (row `all' in Table \ref{tb:in-cross}) does not bring an obvious benefit in general. It is reasonable since BERT is already trained on a general domain.

We also find that IMDb and Yelp do not help each other in sentiment domain. The reason may be that IMDb and Yelp are two sentiment tasks of movie and food. The data distributions have a significant difference.

\subsubsection{Comparisons to Previous Models}
    We compare our model with the following a variety of different methods: CNN-based methods such as Char-level CNN \cite{zhang2015character}, VDCNN \cite{conneau2016very} and DPCNN \cite{johnson2017deep}; RNN-based models such as D-LSTM \cite{yogatama2017generative}, Skim-LSTM \cite{seo2017neural} and hierarchical attention networks \cite{yang2016hierarchical}; feature-based transfer learning methods such as rigion embedding \cite{qiao2018anew} and CoVe \cite{mccann2017learned}; and the language model fine-tuning method (ULMFiT) \cite{howard2018universal}, which is the current state-of-the-art for text classification.

	We implement BERT-Feat through using the feature from BERT model as the input embedding of the biLSTM with self-attention \cite{lin2017structured}. The result of BERT-IDPT-FiT corresponds to the row of `all sentiment', `all question', and `all topic' in Table \ref{tb:in-cross}, and the result of BERT-CDPT-FiT corresponds to the row of `all' in it.
	
	As is shown in Table \ref{tb:comparison}, BERT-Feat performs better than all other baselines except for ULMFiT. In addition to being slightly worse than BERT-Feat on DBpedia dataset, BERT-FiT outperforms BERT-Feat on the other seven datasets.
	Moreover, all of the three further pre-training models are better than BERT-FiT model. Using BERT-Feat as a reference, we calculate the average percentage increase of other BERT-FiT models on each dataset. BERT-IDPT-FiT performs best, with an average error rate reduce by 18.57\%.

\begin{table*}[h!]\small\setlength{\tabcolsep}{6pt}
		\centering
		\begin{tabular}{l c c c c c c c c c }
			\toprule
			Model & IMDb & Yelp P. & Yelp F. & TREC & Yah. A. & AG & DBP & Sogou & Avg. $\Delta$ \\
			\midrule
			Char-level CNN\cite{zhang2015character} & / & 4.88 & 37.95 & / & 28.80 & 9.51 & 1.55 & 3.80$^*$ & / \\
			VDCNN \cite{conneau2016very} & / & 4.28 & 35.28 & / & 26.57 & 8.67 & 1.29 & 3.28 & / \\
			DPCNN \cite{johnson2017deep} & / & 2.64 & 30.58 & / & 23.90 & 6.87 & 0.88 & 3.48$^*$ & / \\
			\midrule	
			D-LSTM \cite{yogatama2017generative} & / & 7.40 & 40.40 & / & 26.30 & 7.90 & 1.30 & 5.10 & / \\
			Standard LSTM \cite{seo2017neural} & 8.90 & / & / & / & / & 6.50 & / & / & / \\
			Skim-LSTM \cite{seo2017neural} & 8.80 & / & / & / & / & 6.40 & / & / & / \\
			HAN \cite{yang2016hierarchical} & / & / & / & / & 24.20 & / & / & / & / \\
			\midrule
			Region Emb. \cite{qiao2018anew} & / & 3.60 & 35.10 & / & 26.30 & 7.20 & 1.10 & 2.40 & / \\
			CoVe \cite{mccann2017learned} & 8.20 & / & / & 4.20 & / & / & / & / & / \\
			ULMFiT \cite{howard2018universal} & 4.60 & 2.16 & 29.98 & 3.60 & / & 5.01 & 0.80 & / & / \\
			\midrule
			BERT-Feat & 6.79 & 2.39 & 30.47 & 4.20 & 22.72 & 5.92 & 0.70 & 2.50 & - \\ 
			BERT-FiT & 5.40 & 2.28 & 30.06 & 2.80 & 22.42 & 5.25 & 0.71 & 2.43 & 9.22\% \\  
			BERT-ITPT-FiT & \textbf{4.37} & 1.92 & 29.42 & 3.20 & 22.38 & \textbf{4.80} & 0.68 & \textbf{1.93} & 16.07\% \\ 
			BERT-IDPT-FiT & 4.88 & \textbf{1.87} & 29.25 & \textbf{2.20} & \textbf{21.86} & 4.88 & \textbf{0.65} & / & \textbf{18.57}\% \\ 
			BERT-CDPT-FiT & 5.18 & 1.97 & \textbf{29.20} & 2.80 & 21.94 & 5.08 & 0.67 & / & 14.38\% \\ 
			\bottomrule
		\end{tabular}
		\caption{\label{tb:comparison} Test error rates (\%) on eight text classification datasets. The results without $^*$ of previous models are the results reported on their papers. / means not reported. $^*$ means the results are from our implementation since the Sogou dataset is different from theirs.  BERT-Feat means ``BERT as features''. BERT-FiT means ``BERT + Fine-Tuning''.
BERT-ITPT-FiT means ``BERT + withIn-Task Pre-Training + Fine-Tuning''. BERT-IDPT-FiT means ``BERT + In-Domain Pre-Training + Fine-Tuning''.  BERT-CDPT-FiT means ``BERT + Cross-Domain Pre-Training + Fine-Tuning''.
		}
	\end{table*}

\subsection{Exp-III: Multi-task Fine-Tuning} \label{sec:exp-mtl}

\begin{table}[h!]\small\setlength{\tabcolsep}{5pt}
	\centering
	\begin{tabular}{l c c c c}
		\toprule
		Method & IMDb & Yelp P. & AG & DBP \\
		\midrule
		BERT-FiT & 5.40 & 2.28 & 5.25 & 0.71 \\
		BERT-MFiT-FiT & 5.36 & 2.19 & 5.20 & 0.68 \\
		\midrule
		BERT-CDPT-FiT & 5.18 & \textbf{1.97} & \textbf{5.08} & \textbf{0.67} \\
		BERT-CDPT-MFiT-FiT & \textbf{4.96} & 2.06 & 5.13 & \textbf{0.67} \\
		\bottomrule
	\end{tabular}
	\caption{\label{tb:multitask}Test error rates (\%) with multi-task fine-tuning.
	}
\end{table}

When there are several datasets for the text classification task, to take full advantage of these available data, we further consider a fine-tuning step with multi-task learning. We use four English text classification datasets (IMDb, Yelp P., AG, and DBP). The dataset Yelp F. is excluded since there is overlap between the test set of Yelp F. and the training set of Yelp P., and two datasets of question domain are also excluded.

We experiment with the official uncased BERT-base weights and the weights further pre-trained on all seven English classification datasets respectively. In order to achieve better classification results for each subtask, after fine-tuning together, we fine-tune the extra steps on the respective datasets with a lower learning rate.

Table \ref{tb:multitask} shows that for multi-task fine-tuning based on BERT, the effect is improved. However, multi-task fine-tuning does not seem to be helpful to BERT-CDPT in Yelp P. and AG. Multi-task fine-tuning and cross-domain pre-training may be alternative methods since the BERT-CDPT model already contains rich domain-specific information, and multi-task learning may not be necessary to improve generalization on related text classification sub-tasks.


\subsection{Exp-IV: Few-Shot Learning}\label{sec:exp-few}

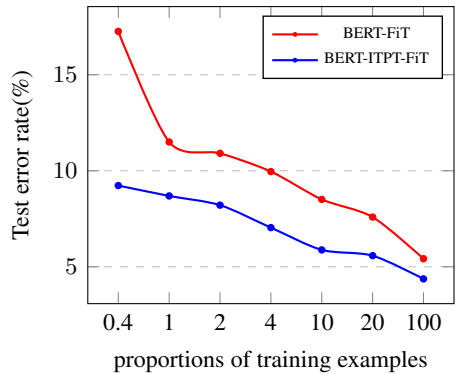
\begin{figure}[h!]
    \centering
    \pgfplotsset{width=0.4\textwidth}
    \begin{tikzpicture}
    \pgfplotstableread{./fewshot.txt}\datatable
      \begin{axis}[
      font=\small,
      xlabel={proportions of training examples},
      xtick=data,
        xticklabels from table={\datatable}{[index] 0},
        xtick distance=1,
        table/x expr=\coordindex,
      ylabel={Test error rate(\%)},
      legend entries={BERT-FiT, BERT-ITPT-FiT},
      mark size=1.0pt,
      ymajorgrids=true,
      grid style=dashed,
      legend pos= north east,
      legend style={font=\tiny,line width=.5pt,mark size=.5pt,
              /tikz/every even column/.append style={column sep=0.5em}},
              smooth,
      ]

      \addplot [red,thick,mark=*] table [x=data, y=BERT] from \datatable;
        \addplot [blue,thick,mark=*] table [x=data, y=Fit] from \datatable;


      \end{axis}
  \end{tikzpicture}
  \caption{Test error rates(\%) on IMDb dataset with different proportions of training examples.}\label{fig:few-show}
  \end{figure}

One of the benefits of the pre-trained model is being able to train a model for downstream tasks within small training data. We evaluate BERT-FiT and BERT-ITPT-FiT on different numbers of training examples. We select a subset of IMDb training data and feed them into BERT-FiT and BERT-ITPT-FiT. We show the result in Figure \ref{fig:few-show}.

This experiment result demonstrates that BERT brings a significant improvement to small size data. Further pre-trained BERT can further boost its performance, which improves the performance from 17.26\% to 9.23\% in error rates with only 0.4\% training data.


\subsection{Exp-V: Further Pre-Training on BERT Large}
In this subsection, we investigate whether the  BERT$_\mathrm{LARGE}$ model has similar findings to  BERT$_\mathrm{BASE}$. We further pre-train Google's pre-trained BERT$_\mathrm{LARGE}$ model\footnote{https://storage.googleapis.com/bert\_models/2018\_10\_18/\\uncased\_L-24\_H-1024\_A-16.zip} on 1 Tesla-V100-PCIE 32G GPU with a batch size of 24, the max sequence length of 128 and 120K training steps. For target task classifier BERT fine-tuning, we set the batch size to 24 and fine-tune BERT$_\mathrm{LARGE}$ on 4 Tesla-V100-PCIE 32G GPUs with the max sequence length of 512.

As shown in Table \ref{tb:comparison_large}, ULMFiT performs better on almost all of the tasks compared to BERT$_\mathrm{BASE}$ but not BERT$_\mathrm{LARGE}$. This changes however with the task-specific further pre-training where even BERT$_\mathrm{BASE}$ outperforms ULMFiT on all tasks. BERT$_\mathrm{LARGE}$ fine-tuning with task-specific further pre-training achieves state-of-the-art results.

\begin{table}[h]\small\setlength{\tabcolsep}{5pt}
		\centering
		\begin{tabular}{l | c c c c c}
			\toprule
			Model & IMDb & Yelp P. & Yelp F. & AG & DBP  \\
			\midrule
			ULMFiT & 4.60 & 2.16 & 29.98 & 5.01 & 0.80 \\
			\midrule
			BERT$_\mathrm{BASE}$ & 5.40 & 2.28 & 30.06 & 5.25 & 0.71 \\
			\qquad + ITPT & 4.37 & 1.92 & 29.42 & 4.80 & 0.68 \\
			BERT$_\mathrm{LARGE}$ & 4.86 & 2.04 & 29.25 & 4.86 & 0.62  \\
			\qquad + ITPT & \textbf{4.21} & \textbf{1.81} & \textbf{28.62} & \textbf{4.66} & \textbf{0.61} \\
			\bottomrule
		\end{tabular}
		\caption{\label{tb:comparison_large} Test error rates (\%) on five text classification datasets.
		}
\end{table}

\section{Conclusion}

In this paper, we conduct extensive experiments to investigate the different approaches to fine-tuning BERT for the text classification task. There are some experimental findings: 1) The top layer of BERT is more useful for text classification; 2) With an appropriate layer-wise decreasing learning rate, BERT can overcome the catastrophic forgetting problem; 3) Within-task and in-domain further pre-training can significantly boost its performance; 4) A preceding multi-task fine-tuning is also helpful to the single-task fine-tuning, but its benefit is smaller than further pre-training; 5) BERT can improve the task with small-size data.

With the above findings, we achieve state-of-the-art performances on eight widely studied text classification datasets. In the future, we will probe more insight of BERT on how it works.

	
	\bibliography{sunnysc_modified}
	\bibliographystyle{acl_natbib}
	\end{CJK*}
\end{document}